\pdfoutput=1

\documentclass[11pt]{article}

\usepackage[final]{acl}

\usepackage{times}
\usepackage{latexsym}

\usepackage[T1]{fontenc}

\usepackage[utf8]{inputenc}

\usepackage{microtype}

\usepackage{inconsolata}

\usepackage{graphicx}
\usepackage{url}
\usepackage{enumitem}
\usepackage{booktabs}
\usepackage{makecell}
\usepackage{multirow}
\usepackage{algorithm}
\usepackage{algpseudocode}
\usepackage{amssymb}
\usepackage{amsmath}
\usepackage{pifont}
\usepackage[most]{tcolorbox}
\usepackage{soul}
\usepackage{xcolor}
\usepackage{ulem}
\usepackage{colortbl}

\usepackage{pgfplotstable}
\usepackage{subfigure}
\usepackage{pgfplots}
\usepackage{tikz}
\usetikzlibrary{arrows.meta}
\usetikzlibrary{decorations}
\usetikzlibrary{pgfplots.groupplots}
\usepgfplotslibrary{fillbetween}
\usetikzlibrary{shapes.multipart}
\usetikzlibrary{shapes.geometric}

\definecolor{ugreen}{cmyk}{1,0,1,0.498}
\definecolor{lyyblue}{cmyk}{0.8278,0.3333,0,0.2941}
\definecolor{lyygreen}{cmyk}{0.6813,0,0.725,0.3725}
\definecolor{lyyred}{cmyk}{0,0.8855,0.8767,0.1098}
\definecolor{dblue}{cmyk}{1,0.5487,0,0.5569}
\definecolor{royalblue}{HTML}{4169e1}

\definecolor{myred}{HTML}{E33222}

\newcommand{\cmark}{\ding{52}}%
\newcommand{\xmark}{\ding{56}}%

\definecolor{gr}{RGB}{0, 146, 0}
\newcommand{\gr}[1]{\textcolor{gr}{#1}}
\newcommand{\red}[1]{\textcolor{myred}{#1}}

\newcommand{\org}[1]{\textcolor{orange}{#1}}
\newcommand{\blue}[1]{\textcolor{blue}{#1}}

\algnewcommand{\LineComment}[1]{\State \(\triangleright\) #1}

\newcommand{\greencheck}{\textcolor{gr}{\ding{51}}}
\newcommand{\redcross}{\textcolor{myred}{\ding{55}}}

\newcommand{\method}{\textsc{OpTune}}

\title{Learning to Reason from Feedback at Test-Time}

\author{Yanyang Li, Michael R. Lyu, Liwei Wang\thanks{Corresponding author.}\\
The Chinese University of Hong Kong \\
\texttt{\{yyli21,lyu,lwwang\}@cse.cuhk.edu.hk}
}

\begin{document}
\maketitle
\begin{abstract}
Solving complex tasks in a single attempt is challenging for large language models (LLMs). Iterative interaction with the environment and feedback is often required to achieve success, making effective feedback utilization a critical topic. Existing approaches either struggle with length generalization or rely on naive retries without leveraging prior information. In this paper, we introduce FTTT, a novel paradigm that formulates feedback utilization as an optimization problem at test time. Additionally, we propose a learnable test-time optimizer, \method{}, to effectively exploit feedback. Experiments on two LLMs across four reasoning datasets demonstrate that FTTT and \method{} achieve superior scalability and performance\footnote{\url{https://github.com/LaVi-Lab/FTTT}}.
\end{abstract}

\section{Introduction}

Leveraging external feedback from interactions with the environment during test time has emerged as a promising approach for large language models (LLMs). This includes applications such as LLM-based agents~\cite{DBLP:conf/iclr/YaoZYDSN023,DBLP:conf/nips/ShinnCGNY23} and, more recently, test-time scaling~\cite{wu2024inferencescalinglawsempirical,DBLP:journals/corr/abs-2408-03314,liu20251bllmsurpass405b}. Such methods further enhance the potential of LLMs to solve challenging tasks, e.g., Olympiad-level math problems~\cite{guan2025rstarmathsmallllmsmaster} and competitive programming~\cite{openai2025competitiveprogramminglargereasoning}.

Significant progress in this area typically falls into two categories~\cite{DBLP:journals/corr/abs-2408-03314}, as illustrated in Figure~\ref{fig:compare}: sequential revision and parallel sampling. Sequential revision methods~\cite{DBLP:conf/nips/ShinnCGNY23,DBLP:conf/nips/MadaanTGHGW0DPY23} incorporate previous attempts into the LLM's context, while parallel sampling methods~\cite{DBLP:conf/nips/XieKZZKHX23,DBLP:journals/corr/abs-2407-21787} generate new attempts independently of prior failures.
However, both approaches have notable limitations. Sequential revision is computationally expensive due to long context lengths and faces challenges~\cite{muennighoff2025s1simpletesttimescaling}, such as position bias~\cite{DBLP:journals/tacl/LiuLHPBPL24} and attention noise~\cite{ye2024differentialtransformer}. In contrast, parallel sampling, while efficient, fails to learn from previous errors~\cite{DBLP:journals/corr/abs-2407-21787}.
Unlike these paradigms, human reasoning follows a different pattern: humans store recent experiences in ``fast weights''~\cite{DBLP:journals/neco/Schmidhuber92,DBLP:conf/nips/BaHMLI16,pmlr-v139-schlag21a}, enabling them to neither revisit past errors explicitly nor start each attempt without any prior knowledge. Recent research suggests that the weights of neural networks could serve as a natural memory mechanism during test time~\cite{DBLP:journals/corr/abs-2401-11504}.




Building on these observations, we propose a novel paradigm that leverages Test-Time Training (TTT)~\cite{DBLP:conf/icml/SunWLMEH20,DBLP:journals/corr/abs-2310-13807} to store past experiences in model weights rather than in the context. This approach bridges the gap between sequential revision and parallel sampling by indirectly incorporating knowledge into the LLM without disrupting in-context reasoning. Specifically, we introduce \textbf{F}eedback-based \textbf{T}est-\textbf{T}ime \textbf{T}raining (FTTT), which employs a carefully designed TTT task based on feedback enriched with self-reflection. We demonstrate that FTTT improves test-time computation scalability on two mathematical reasoning and two code generation datasets, using \texttt{Llama-3.1-8B-Instruct}~\cite{DBLP:journals/corr/abs-2407-21783} and \texttt{Mistral-7B-Instruct-v0.3}~\cite{DBLP:journals/corr/abs-2310-06825}.

\begin{figure*}[t!]
    \newdimen\base
    \base=0.7cm

    \tikzstyle{textnode} = [rectangle,font=\scriptsize,draw=black,inner sep=0pt,outer sep=0pt,minimum width=2\base,minimum height=\base,rounded corners=2pt]

    \hspace*{\fill}
    \subfigure[Sequential Revision]
    {
        \centering
        \begin{tikzpicture}
        
        \node[textnode, fill=lyyblue!40!white] (question) at (0,0) {Question};
        \node[textnode, anchor=north, fill=red!30!white] (answer1) at ([yshift=-0.5\base]question.south) {Attempt \#1};
        \node[textnode, anchor=north, fill=red!30!white] (answer2) at ([yshift=-0.5\base]answer1.south) {Attempt \#2};
        \node[textnode, anchor=north, fill=red!30!white] (answer3) at ([yshift=-0.5\base]answer2.south) {Attempt \#3};
        \node[textnode, anchor=north, fill=ugreen!30!white] (answer4) at ([yshift=-0.5\base]answer3.south) {Attempt \#4};
        
        \draw[-latex] (question.east) to[out=0, in=0, looseness=3] ([xshift=\base, yshift=-0.25\base]question.south) to ([xshift=-\base, yshift=0.25\base]answer1.north) to[out=180, in=180, looseness=3] (answer1.west);
        \draw[-latex] (answer1.east) to[out=0, in=0, looseness=3] ([xshift=\base, yshift=-0.25\base]answer1.south) to ([xshift=-\base, yshift=0.25\base]answer2.north) to[out=180, in=180, looseness=3] (answer2.west);
        \draw[-latex] (answer2.east) to[out=0, in=0, looseness=3] ([xshift=\base, yshift=-0.25\base]answer2.south) to ([xshift=-\base, yshift=0.25\base]answer3.north) to[out=180, in=180, looseness=3] (answer3.west);
        \draw[-latex] (answer3.east) to[out=0, in=0, looseness=3] ([xshift=\base, yshift=-0.25\base]answer3.south) to ([xshift=-\base, yshift=0.25\base]answer4.north) to[out=180, in=180, looseness=3] (answer4.west);
        
        \end{tikzpicture}
        \label{fig:revision}
    }
    \hfill
    \subfigure[Parallel Sampling]
    {
        \centering
        \begin{tikzpicture}
        
        \node[textnode, fill=lyyblue!40!white] (question) at (0,0) {Question};
        \node[textnode, anchor=west, fill=red!30!white] (answer1) at ([xshift=2\base, yshift=3\base]question.east) {Attempt \#1};
        \node[textnode, anchor=west, fill=ugreen!30!white] (answer2) at ([xshift=2\base, yshift=\base]question.east) {Attempt \#2};
        \node[textnode, anchor=west, fill=red!30!white] (answer3) at ([xshift=2\base, yshift=-\base]question.east) {Attempt \#3};
        \node[textnode, anchor=west, fill=red!30!white] (answer4) at ([xshift=2\base, yshift=-3\base]question.east) {Attempt \#4};

        \draw[-latex] (question.east) to[out=0, in=180] (answer1.west);
        \draw[-latex] (question.east) to[out=0, in=180] (answer2.west);
        \draw[-latex] (question.east) to[out=0, in=180] (answer3.west);
        \draw[-latex] (question.east) to[out=0, in=180] (answer4.west);
        
        \end{tikzpicture}
      \label{fig:search}
    }
    \hfill
    \subfigure[Feedback-based Test-Time Training]
    {
        \centering
        \begin{tikzpicture}
        
        \node[textnode, fill=lyyblue!40!white] (question) at (0,0) {Question};
        \node[textnode, anchor=west, fill=red!30!white] (answer1) at ([xshift=2\base, yshift=3\base]question.east) {Attempt \#1};
        \node[textnode, anchor=west, fill=red!30!white] (answer2) at ([xshift=2\base, yshift=\base]question.east) {Attempt \#2};
        \node[textnode, anchor=west, fill=red!30!white] (answer3) at ([xshift=2\base, yshift=-\base]question.east) {Attempt \#3};
        \node[textnode, anchor=west, fill=ugreen!30!white] (answer4) at ([xshift=2\base, yshift=-3\base]question.east) {Attempt \#4};

        \draw[-latex] (question.east) to[out=0, in=180] (answer1.west);
        \draw[-latex] (question.east) to[out=0, in=180] (answer2.west);
        \draw[-latex] (question.east) to[out=0, in=180] (answer3.west);
        \draw[-latex] (question.east) to[out=0, in=180] (answer4.west);
        
        \draw[-latex, densely dashed] (answer1.east) to[out=0, in=0, looseness=2] ([xshift=\base, yshift=-0.5\base]answer1.south) to ([xshift=-\base, yshift=0.5\base]answer2.north) to[out=180, in=180, looseness=2] (answer2.west);
        \draw[-latex, densely dashed] (answer2.east) to[out=0, in=0, looseness=2] ([xshift=\base, yshift=-0.5\base]answer2.south) to ([xshift=-\base, yshift=0.5\base]answer3.north) to[out=180, in=180, looseness=2] (answer3.west);
        \draw[-latex, densely dashed] (answer3.east) to[out=0, in=0, looseness=2] ([xshift=\base, yshift=-0.5\base]answer3.south) to ([xshift=-\base, yshift=0.5\base]answer4.north) to[out=180, in=180, looseness=2] (answer4.west);
        
        \end{tikzpicture}
      \label{fig:ttt}
    }
    \hspace*{\fill}
    \caption{Comparison between sequential revision, parallel sampling, and feedback-based test-time training. {\protect\tikz \protect\draw[color=black, fill=red!30!white] plot[mark=square*, mark options={scale=1.4}] (0,0);} is the failed attempt and {\protect\tikz \protect\draw[color=black, fill=ugreen!30!white] plot[mark=square*, mark options={scale=1.4}] (0,0);} is the successful attempt. {\protect\tikz {\protect\draw[-latex, thick, color=black] (0,0.5) -- (0.5,0.5);\protect\draw[opacity=0] (0,0.4) -- (0.5,0.4);}} indicates the LLM generation with the input on the left of the arrow and the output on the right. {\protect\tikz {\protect\draw[-latex, thick, densely dashed, color=black] (0,0.5) -- (0.5,0.5);\protect\draw[opacity=0] (0,0.4) -- (0.5,0.4);}} denotes the LLM training, where the left of the arrow is the training data.}
    \label{fig:compare}
\end{figure*}

Inspired by advancements in learning to optimize~\cite{DBLP:journals/jmlr/ChenCCH0WY22}, we explore training learnable test-time optimizers to yield Pareto-optimal cost-performance tradeoffs. Our proposed learnable optimizer, \method{}, is a lightweight neural network that predicts weight updates from the gradients of the previous attempt. Unlike traditional parameter-efficient fine-tuning (PEFT) methods, \method{} works on the gradient rather than the activation space. Experiments on three reasoning datasets and two different LLMs demonstrate the effectiveness of \method{}, outperforming five widely used PEFT baselines.


\section{Feedback-based Test-Time Training}
\label{sec:ttt}

\subsection{The Test-Time Training Task}

The problem of exploiting test-time feedback is as~\cite{DBLP:conf/nips/ShinnCGNY23}: given a question $Q$, a model $M$ attempts to solve $Q$ within a budget of $N$ attempts. A verifier $V$ evaluates each attempt, such as the $n$-th attempt $A_n$, and provides feedback $V(A_n)$. This work focuses on binary verifiers, which determine whether $A_n$ is correct. These verifiers are well-established, rule-based systems that are both cost-effective and efficient to evaluate.

When the model generates attempts sequentially, our goal is to enable $M$ to learn from previous attempts to improve subsequent ones. To achieve this, we frame learning from previous attempts as a training problem: at each step $n$, we optimize $M$ using $Q$, $A_n$, and $V(A_n)$, aiming for $M$ to generate a better $A_{n+1}$. This way internalizes the past attempts into weights for efficient inference of $A_{n+1}$. As a result, the sequence of attempts can be viewed as an $N$-step optimization process.

A key challenge is designing an effective supervised task using $Q$, $A_n$, and $V(A_n)$ to improve the model's ability to solve $Q$. We build on the intuition that \textit{a model capable of judging the correctness of a solution should also be able to solve the question itself}. Concretely, given $Q$ and $A_n$, we train $M$ to predict verbal feedback $F$ that aligns with $V(A_n)$. This leads to our FTTT loss:
\begin{equation}
\scalebox{0.8}{\ensuremath{
    \mathcal{L}_\mathrm{FTTT}\left(Q, A_n\right)=-\frac{1}{l_0}\log M_{n-1}\left(F\mid Q, A_n\right)
}}
    \label{eqn:ttt-loss}
\end{equation}
where $l_0$ is the length of $F$ and $M_0$ denotes the raw LLM. In this work, $F$ is set to ``Your answer is incorrect.'' when $V\left( A_n\right)$ implies an incorrect $A_n$.

\subsection{Self-Reflected Feedback}

Since we are working with a binary verifier, the learning signal is limited at each interaction. Previous research suggests that LLMs can self-correct errors when provided with external signals~\cite{DBLP:conf/iclr/0009CMZYSZ24}. Inspired by this, we aim to enhance the learning signal by leveraging the model to generate silver-standard training labels via self-reflection.

We first sample a reflection $R_n$ from the model given $Q$, $A_n$, $F$ and the instruction $P$ (see Appendix~\ref{sec:prompt} for details of $P$):
\begin{equation}
    R_n\sim M_0\left(R\mid Q,A_n,F,P\right)
\end{equation}
In practice, we use $M_0$ to generate $R_n$ to mitigate the risk of degraded self-reflection ability after training.
The auxiliary loss is then defined as:
\begin{equation}
\scalebox{0.8}{\ensuremath{
    \mathcal{L}_\mathrm{aux}\left(Q, A_n, R_n\right)=-\frac{1}{l_n} \log M_{n-1}\left(R_n\mid Q, A_n, F\right)
}}
    \label{eqn:reflect-loss}
\end{equation}
where $l_n$ is the length of $R_n$. Eq.~\ref{eqn:reflect-loss} can be interpreted as a sequence-level distillation loss~\cite{kim-rush-2016-sequence}, where knowledge from the raw model $M_0$ is distilled into the trained model $M_{n-1}$ to prevent overfitting.
Finally, the overall loss is as:
\begin{equation}
    \mathcal{L}_\mathrm{final}=\mathcal{L}_\mathrm{FTTT} + \mathcal{L}_\mathrm{aux}
\end{equation}
Below is a training example with self-reflection, where underlined sentences are the training target:

\patchcmd{\quote}{\rightmargin}{\leftmargin 15pt \rightmargin}{}{}
\begin{quote}
\small 
\begin{tcolorbox}[breakable, colback=white, colbacktitle=blue!5!white, colframe=black, boxrule=1pt, title={\textcolor{black}{\textbf{Training Example with Self-Reflection}}}]
\textbf{User:} Solve the following math problem $\ldots$\\
\textbf{Assistant:} $\ldots$ the final answer is: $\ldots$\\
\textbf{User:} \ul{Your answer is incorrect.} \uwave{Here is the summary of the mistakes in the previous solution $\ldots$}
\end{tcolorbox}
\end{quote}
\patchcmd{\quote}{\rightmargin}{\leftmargin 26pt \rightmargin}{}{}
The \ul{underlined} sentence corresponds to $F$ in Eq.~\ref{eqn:ttt-loss} and the \uwave{wave-underlined} sentence represents $R_n$ for Eq.~\ref{eqn:reflect-loss}.
Algorithm~\ref{alg:ttt} summarizes our FTTT.

\begin{algorithm}[t!]
    \caption{\textsc{Feedback-based TTT}}
    \label{alg:ttt}
    \begin{algorithmic}[1]
        \Require The question $Q$, the model $M_0$, the verifier $V$, the budget $N$, the verbal feedback $F$, the instruction $P$
        \State $n \leftarrow 1$
        \While{$n \le N$}
            \State $A_n\sim M_{n-1}\left(A\mid Q\right)$ 
            \If{$V\left(A_n\right)$ is passed}
                \State \Return $A_n$
            \Else
                \State Compute the loss $\mathcal{L}$ using Eq.~\ref{eqn:ttt-loss} 
                \If{enable self-reflection}
                    \State $R_n\sim M_0\left(R\mid Q,A_n,F,P\right)$
                    \State Compute the loss $\mathcal{L}_\mathrm{aux}$ using Eq.~\ref{eqn:reflect-loss} 
                    \State $\mathcal{L} \leftarrow \mathcal{L} + \mathcal{L}_\mathrm{aux}$
                \EndIf
                \State Update $M_{n-1}$ using $\mathcal{L}$ to get $M_{n}$
            \EndIf
            \State $n \leftarrow n + 1$
        \EndWhile
        \State \Return $A_{N}$
    \end{algorithmic}
\end{algorithm}

\paragraph{Discussion.}

As shown in Table~\ref{tab:compare}, FTTT combines the strengths of sequential revision and parallel sampling. Like sequential revision, it leverages memory (i.e., weights) to store past failed attempts, while avoiding the length generalization issues as in parallel sampling. Figure~\ref{fig:compare} highlights the advantages of FTTT from a probabilistic graphical model perspective, where both sequential revision and parallel sampling are special cases of FTTT with additional independence assumptions. In particular, sequential revision assumes that attempts form a Markov process, whereas parallel sampling treats each attempt as independent.

In terms of efficiency, FTTT is highly optimized. Since we only update $M$ once at each attempt in practice, FTTT requires only one additional backward propagation computed in parallel for all tokens in one attempt, whose cost is negligible. The dominant overhead of FTTT is generating $R_n$. However, FTTT is still much faster than sequential revision (see Section~\ref{sec:result1}), as $R_n$ is typically short.

Moreover, FTTT closely resembles vanilla TTT~\cite{DBLP:conf/icml/SunWLMEH20}, but with additional inputs beyond $Q$, such as $A_n$, $V\left(A_n\right)$, and $R_n$, as defined in the problem. Consequently, it inherits the same convergence guarantees as TTT.

\begin{table}
\centering
\resizebox{\linewidth}{!}{
\setlength{\tabcolsep}{5pt}
\begin{tabular}[t]{l|ccc}
\toprule
 \makecell[c]{\textbf{Method}} & \makecell[c]{\textbf{Self}\\\textbf{Reflection}} & \makecell[c]{\textbf{Memory}} & \makecell[c]{\textbf{Length}\\\textbf{Generalization}} \\
\midrule
Revision~\cite{DBLP:journals/corr/abs-2408-03314} & \redcross & \greencheck & \redcross \\
Self-Refine~\cite{DBLP:conf/nips/MadaanTGHGW0DPY23} & \greencheck & \greencheck & \redcross \\
Best-of-N~\cite{DBLP:journals/corr/abs-2407-21787} & \redcross & \redcross & \greencheck \\
Beam Search~\cite{ow1988filtered} & \redcross & \redcross & \greencheck \\
Guided Beam Search~\cite{DBLP:conf/nips/XieKZZKHX23} & \greencheck & \redcross & \greencheck \\
\midrule
\textbf{FTTT (ours)} & \greencheck & \greencheck & \greencheck \\
\bottomrule
\end{tabular}
}
\caption{Comparing the advantages and drawbacks of FTTT and related works.}
\label{tab:compare}
\end{table}

\section{A Learnable Test-Time Optimizer}

\subsection{The Learning to Optimize Problem}

Although FTTT achieves success (see Section~\ref{sec:result1}), it simply applies the gradients of the feedback received so far to update the weights. This raises the question: can we design a better test-time optimizer that more effectively exploits feedback?

Motivated by learning to optimize~\cite{DBLP:journals/jmlr/ChenCCH0WY22}, we adopt a neural network as the test-time optimizer. Concretely, this learnable test-time optimizer is formulated as $f_\theta\left(Q, \left\{A_i, V\left(A_i\right)\right\}^n_{i=1}\right)$, which predicts updates for all LLM weights based on the previous $n$ attempts, and $\theta$ is the optimizer parameter. However, this direct formulation leads to prohibitively large networks due to high-dimensional input and output spaces. For a maximum number of $m$ tokens per attempt and an $l$-layer LLM, the input space grows to $n \times m \times l$, even when updating only a scalar weight at each layer (we exclude the token count of $Q$, as it is significantly smaller than $m$). Since updates for all weight matrices across all layers are predicted jointly, the dimensionality of the input and output spaces becomes unmanageable. We therefore simplify $f_\theta$ by introducing the following assumptions:
\begin{enumerate}[noitemsep, nolistsep, label={(\bfseries A\arabic*):},left=\parindent]
    \item \textbf{Markov Property}: The latest attempt captures all relevant information from previous attempts.
    \item \textbf{Independent Update}: The optimizer predicts updates for each weight matrix independently, similar to conventional optimizers~\cite{DBLP:journals/corr/KingmaB14}.
\end{enumerate}

\textbf{A1} eliminates the dependency on $n$ and \textbf{A2} enables updates to be predicted independently for each weight, significantly reducing the size of the output space. The learnable test-time optimizer now becomes $\tilde{\nabla}_{W_i}=f_{\theta_{W_i}}\left(Q, A_n, V\left(A_n\right)\right)$, where it predicts the update for the weight $W_i$ in the $i$-th layer based on $Q$ and the latest attempt $A_n$.

To train all $f$, we define the following loss:
\begin{equation}
\scalebox{0.8}{\ensuremath{
    \mathcal{L}_\mathrm{meta}=-\log M\left(\hat{A} \mid Q, \left\{W+\tilde{\nabla}_{W}\mid\forall \ W \in \mathcal{W}\right\}\right)
}}
    \label{eqn:meta}
\end{equation}
where $\hat{A}$ is the ground-truth for $Q$ and $\mathcal{W}$ is the set of LLM weights. Here we consider only a one-step optimization problem, as a multi-step version involves the non-differentiable generation of $A$. Eq.~\ref{eqn:meta} encourages $f$ to predict updates that increase the likelihood of generating the correct answer after applying these updates.

\subsection{A Parameter-Efficient Architecture in The Gradient Space}

Given the limited learning signal at test time, we design the learnable optimizer to be parameter-efficient to alleviate overfitting. However, the input and output spaces of $f_{\theta_{W_i}}\left(Q, A_n, V\left(A_n\right)\right)$ are large due to their lengths, making even a simple linear projection parameter-intensive. Additionally, $V\left(A_n\right)$ may be heterogeneous to $Q$ and $A_n$, e.g., a scalar, posing challenges for modeling.

Inspired by the success of FTTT in Section~\ref{sec:ttt} and recent works~\cite{DBLP:conf/iclr/MitchellLBFM22,DBLP:journals/corr/abs-2401-11504}, we propose a parameter-efficient architecture in the gradient space as the learnable optimizer.

\paragraph{Gradient-based Input Compression.}

Instead of directly inputting $Q$, $A_n$, and $V\left(A_n\right)$, we first project them into the gradient space, since recent work suggests that long context can be effectively compressed by gradients~\cite{DBLP:journals/corr/abs-2401-11504}. This way reduces the token count $m$ in $A_n$ to a constant and unifies the spaces of $Q$, $A_n$, and $V\left(A_n\right)$ to ease the modeling. Besides, the gradient encodes the knowledge about $W_i$, enabling efficient modeling of $f_{\theta_{W_i}}$, which no longer needs to learn such knowledge with massive resources as in the black-box optimization~\cite{DBLP:conf/icml/ChenHCDLBF17}. To compress $Q$ and $A_n$, we use the next token prediction loss, while for $V\left(A_n\right)$, we include $\mathcal{L}_\mathrm{FTTT}$ in Eq.~\ref{eqn:ttt-loss}. The final loss for compressing the optimizer input is:
\begin{equation}
    \mathcal{L}_\mathrm{compress}=-\frac{1}{m}\log M\left(A_n\mid Q\right) + \mathcal{L}_\mathrm{FTTT}
\end{equation}
The input of $f_{\theta_{W_i}}$ to predict the update of $W_i$ now is the gradient $\nabla_{W_i}$ of $\mathcal{L}_\mathrm{compress}$ w.r.t. $W_i$. Consequently, $f_{\theta_{W_i}}$ receives a fixed-size tensor as input rather than a variable-length sequence.

\paragraph{Gradient Decomposition.}

Although $f_{\theta_{W_i}}$ operates on a smaller space after compression, the dimensionality of the gradient space remains large for direct processing. We utilize the observation that $\nabla_{W_i}\in \mathbb{R}^{d\times d}$ (assuming $W_i\in \mathbb{R}^{d\times d}$) can be decomposed into two vectors to further reduce the dimensionalities~\cite{DBLP:conf/iclr/MitchellLBFM22}: the input to a linear projection with weight $W_i$, $u_i\in \mathbb{R}^{d \times 1}$, and the gradient of $\mathcal{L}_\mathrm{compress}$ w.r.t. the output of the same linear projection, $\delta_{i+1}\in \mathbb{R}^{d \times 1}$. In this framework, $f_{\theta_{W_i}}$ takes the decomposed $u_i$ and $\delta_{i+1}$ as its input and predicts $\tilde{u}_i$ and $\tilde{\delta}_{i+1}$. The update is then reconstructed as $\tilde{\nabla}_{W_i}=\tilde{\delta}_{i+1}\tilde{u}_i^T$. This approach reduces the dimension from $d^2$ to $2d$.

\begin{figure}[t!]
    \newdimen\base
    \base=0.5cm
    \newdimen\xsep
    \xsep=0.3cm

    \tikzstyle{common} = [font=\scriptsize]
    \tikzstyle{split2} = [rectangle split, rectangle split parts=2, draw=black, inner sep=0pt, outer sep=0pt, minimum width=\base, minimum height=\base, inner ysep=0.5\base, rounded corners=2pt]
    
    \centering
    \begin{tikzpicture}
    
    \node[common, inner sep=0pt] (input) at (0,0) {$\nabla_{W_i}$};
    \node[common, split2, anchor=west, fill=lyyblue!40!white] (decomposein) at ([xshift=\xsep]input.east) {\rotatebox{-90}{$u_i$} \nodepart{two} \rotatebox{-90}{$\delta_{i+1}$}};
    \node[common, circle, draw=black, inner sep=1pt, anchor=west, fill=orange!30!white] (norm) at ([xshift=\xsep]decomposein.east) {norm};
    \node[common, split2, anchor=west, fill=lyyblue!40!white] (normout) at ([xshift=\xsep]norm.east) {\rotatebox{-90}{$\bar{u}_i$} \nodepart{two} \rotatebox{-90}{$\bar{\delta}_{i+1}$}};
    \node[common, trapezium, trapezium angle=30, rotate=-90, draw=black, anchor=south, fill=ugreen!30!white, minimum height=\base, minimum width=3\base, inner sep=0pt, outer sep=0pt, trapezium stretches, text width=2.5\base, align=center] (down) at ([xshift=\xsep]normout.east) {Down};
    \node[common, rectangle, draw=black, anchor=west, fill=red!30!white, minimum height=2.5\base] (hidden) at ([xshift=\xsep]down.north) {\rotatebox{-90}{Dropout}};
    \node[common, trapezium, trapezium angle=30, rotate=90, draw=black, anchor=north, fill=ugreen!30!white, minimum height=\base, minimum width=3\base, inner sep=0pt, outer sep=0pt, trapezium stretches, text width=2.5\base, align=center] (up) at ([xshift=\xsep]hidden.east) {Up};
    \node[common, anchor=west, inner sep=0pt] (residual) at ([xshift=0.7\xsep]up.south) {$\bigoplus$};
    \node[common, split2, anchor=west, fill=lyyblue!40!white] (decomposeout) at ([xshift=0.7\xsep]residual.east) {\rotatebox{-90}{$\tilde{u}_i$} \nodepart{two} \rotatebox{-90}{$\tilde{\delta}_{i+1}$}};
    \node[common, inner sep=0pt, anchor=west] (out) at ([xshift=\xsep]decomposeout.east) {$\tilde{\nabla}_{W_i}$};

    \draw[-latex] (input) to (decomposein);
    \draw[-latex] (decomposein) to (norm);
    \draw[-latex] (norm) to (normout);
    \draw[-latex] (normout) to (down);
    \draw[-latex] (down) to (hidden);
    \draw[-latex] (hidden) to (up);
    \draw[-latex] (up) to (residual);
    \draw[-latex] (residual) to (decomposeout);
    \draw[-latex] (decomposeout) to (out);

    \draw[-latex] (normout.north) |- ($(normout.north) + (0,0.5\base)$) -| (residual.north);
    
    \end{tikzpicture}
    \caption{The model architecture of \method{}.}
    \label{fig:arch}
\end{figure}

\paragraph{Model Architecture.}

The architecture of $f_{\theta_{W_i}}\left(u_i, \delta_{i+1}\right)$, named \method{}, is shown in Figure~\ref{fig:arch} and defined as follows:
\begin{align}
    \left[\bar{u}_i, \bar{\delta}_{i+1}\right]&=\mathrm{Norm}\left(\left[u_i, \delta_{i+1}\right]\right)\\
    h_i&=\theta_2\mathrm{Dropout}\left(\theta_1\left[\bar{u}_i, \bar{\delta}_{i+1}\right]\right)\\
    \left[\tilde{u}_i, \tilde{\delta}_{i+1}\right]&=h_i + \left[\bar{u}_i, \bar{\delta}_{i+1}\right]
\end{align}
where $\theta_1 \in \mathbb{R}^{r \times 2d}$ and $\theta_2 \in \mathbb{R}^{2d \times r}$ are the optimizer parameters with $r \ll d$. $[\cdot]$ denotes the vector concatenation. $\mathrm{Norm}$ normalizes $u_i$ and $\delta_{i+1}$ to have zero mean and unit variance separately. $\mathrm{Dropout}$ is the dropout regularization~\cite{DBLP:journals/jmlr/SrivastavaHKSS14}. In practice, $\theta_1$ and $\theta_2$ are shared across all weights with the same shape. We design \method{} this way as rank reduction and dropout help mitigate the overfitting risk, which is severe in our scenario due to the sparse, binary feedback. \method{} is similar to the Bottleneck Adapter~\cite{DBLP:conf/icml/HoulsbyGJMLGAG19}, with the key difference that its input is gradients and its output is the weight update. As such, \method{} can also be regarded as a specialized PEFT technique tailored to reasoning.

\begin{table*}
\centering
\resizebox{\linewidth}{!}{
\begin{tabular}[t]{l|llll|c}
\toprule
 \makecell[c]{\textbf{Method}} & \makecell[c]{\textbf{MATH}} & \makecell[c]{\textbf{GSM8K}} & \makecell[c]{\textbf{MBPP}} & \makecell[c]{\textbf{HumanEval}} & \makecell[c]{\textbf{Avg.}} \\
\midrule
\multicolumn{6}{c}{\texttt{Llama-3.1-8B-Instruct}} \\
\midrule
Revision~\cite{DBLP:journals/corr/abs-2408-03314} & 0.2960 & 0.4570 & 0.2991 & 0.3889 & 0.3603 \\
Beam Search~\cite{ow1988filtered} & 0.2560 & 0.4842 & 0.1869 & 0.2407 & 0.2920 \\
Self-Consistency~\cite{DBLP:conf/iclr/0002WSLCNCZ23} & 0.3000$_{0.0057}$ & 0.4525$_{0.0111}$ & 0.1371$_{0.0180}$ & 0.1235$_{0.0531}$ & 0.2533 \\
Self-Refine~\cite{DBLP:conf/nips/MadaanTGHGW0DPY23} & 0.4693$_{0.0207}$ & 0.7828$_{0.0064}$ & 0.2305$_{0.0088}$ & 0.2963$_{0.0800}$ & 0.4447 \\
Best-of-N~\cite{DBLP:journals/corr/abs-2407-21787} & 0.6427$_{0.0154}$ & \underline{0.8069}$_{0.0043}$ & \underline{0.5452}$_{0.0154}$ & 0.6728$_{0.0087}$ & 0.6669 \\
\midrule
FTTT & \underline{0.6707}$_{0.0222}$ & \textbf{0.8100}$_{0.0037}$ & \textbf{0.5607}$_{0.0212}$ & \textbf{0.6852}$_{0.0302}$ & 0.6817 \\
+ w/o Self-Reflected Feedback & \textbf{0.6720}$_{0.0113}$ & 0.8054$_{0.0133}$ & 0.5405$_{0.0117}$ & \underline{0.6790}$_{0.0087}$ & 0.6742 \\
\toprule
\multicolumn{6}{c}{\texttt{Mistral-7B-Instruct-v0.3}} \\
\midrule
Revision~\cite{DBLP:journals/corr/abs-2408-03314} & 0.0497 & 0.1686 & 0.1351 & 0.1000 & 0.1134 \\
Beam Search~\cite{ow1988filtered} & 0.1783 & 0.4537 & 0.1318 & 0.1600 & 0.2310 \\
Self-Consistency~\cite{DBLP:conf/iclr/0002WSLCNCZ23} & 0.1618$_{0.0011}$ & 0.3549$_{0.0052}$ & 0.1243$_{0.0135}$ & 0.1033$_{0.0236}$ & 0.1861 \\
Self-Refine~\cite{DBLP:conf/nips/MadaanTGHGW0DPY23} & 0.1287$_{0.0139}$ & 0.3312$_{0.0078}$ & 0.2347$_{0.0091}$ & 0.3533$_{0.0309}$ & 0.2620 \\
Best-of-N~\cite{DBLP:journals/corr/abs-2407-21787} & 0.4688$_{0.0138}$ & 0.7807$_{0.0055}$ & \textbf{0.4962}$_{0.0171}$ & 0.6500$_{0.0141}$ & 0.5989 \\
\midrule
FTTT & \underline{0.4733}$_{0.0087}$ & \underline{0.7820}$_{0.0045}$ & \textbf{0.4962}$_{0.0015}$ & \underline{0.6633}$_{0.0125}$ & 0.6037 \\
+ w/o Self-Reflected Feedback & \textbf{0.4876}$_{0.0133}$ & \textbf{0.7858}$_{0.0021}$ & \underline{0.4941}$_{0.0040}$ & \textbf{0.6833}$_{0.0205}$ & 0.6127 \\
\bottomrule
\end{tabular}
}
\caption{Experimental results on four datasets with a budget of 32. For stochastic algorithms, we report the mean of three runs with different random seeds and standard deviation in the subscript. \textbf{Bold} entries are the best results, and \underline{underlined} entries are the second-best results.}
\label{tab:pilot}
\end{table*}

\section{Experiments}
\label{sec:exp}

\subsection{Setup}

\paragraph{Datasets.}

We evaluate both baselines and our method on math and coding reasoning tasks:
(a) Mathematical reasoning: MATH~\cite{DBLP:conf/nips/HendrycksBKABTS21} and GSM8K~\cite{DBLP:journals/corr/abs-2110-14168}, using the test split from~\citet{DBLP:conf/iclr/LightmanKBEBLLS24} for MATH.
(b) Code generation: MBPP~\cite{DBLP:journals/corr/abs-2108-07732} and HumanEval~\cite{DBLP:journals/corr/abs-2107-03374}.
For all datasets, we report results on subsets where models fail with greedy decoding. We use Exact Match as the evaluation metric as well as the verifier for math tasks and Pass@1 for code.

\paragraph{Models.}

We conduct experiments with \texttt{Llama-3.1-8B-Instruct}~\cite{DBLP:journals/corr/abs-2407-21783} and \texttt{Mistral-7B-Instruct-v0.3}~\cite{DBLP:journals/corr/abs-2310-06825}.
We evaluate both models with zero-shot prompting and follow the official instructions when evaluating the model on each dataset\footnote{\url{https://huggingface.co/datasets/meta-llama/Llama-3.1-8B-Instruct-evals}}.

\paragraph{Baselines.}

We compare FTTT with the following test-time scaling methods:
\begin{itemize}[noitemsep, nolistsep]
    \item \textbf{Beam Search}~\cite{ow1988filtered} is a popular search algorithm that approximates the most confident model prediction.
    \item \textbf{Self-Consistency}~\cite{DBLP:conf/iclr/0002WSLCNCZ23} samples multiple predictions and selects the most frequent answer.
    \item \textbf{Best-of-N}~\cite{DBLP:journals/corr/abs-2407-21787} samples $N$ predictions independently and picks the best one based on external feedback.
    \item \textbf{Revision}~\cite{DBLP:journals/corr/abs-2408-03314} iteratively refines answers by conditioning the model on previous attempts. We pick the best answer according to feedback.
    \item \textbf{Self-Refine}~\cite{DBLP:conf/nips/MadaanTGHGW0DPY23} alternates between self-critique and refinement. We select the best solution based on feedback.
\end{itemize}

\input{figures/scale}

For sampling-based methods, we use nucleus sampling~\cite{DBLP:conf/iclr/HoltzmanBDFC20} with a temperature of 0.6 and $p=0.95$, following~\citet{DBLP:journals/corr/abs-2407-21787}. All methods are allocated a budget of 32. For FTTT, we fine-tune the model with LoRA~\cite{DBLP:conf/iclr/HuSWALWWC22}, using a rank of 4 and a dropout ratio of 0.05. We use the Adam optimizer~\cite{DBLP:journals/corr/KingmaB14} with a learning rate of 1e-5, except for \texttt{Mistral-7B-Instruct-v0.3} in coding tasks, where we use 2e-5.

As \method{} is a specialized PEFT method, we compare it with the following PEFT approaches: \textbf{Adapter}~\cite{DBLP:conf/icml/HoulsbyGJMLGAG19}, \textbf{(IA)$^3$}~\cite{DBLP:conf/nips/LiuTMMHBR22}, \textbf{LoRA}~\cite{DBLP:conf/iclr/HuSWALWWC22} and \textbf{LN-Tuning}~\cite{DBLP:conf/iclr/ZhaoT0MX24}. We also include \textbf{full fine-tuning} that updates all LLM weights. For \method{}, $r=16$ and the dropout ratio is 0.1. We only apply \method{} to the query and value projections in the last two layers of the LLM. Detailed configurations are in Appendix~\ref{app:setting}.




\subsection{Test-Time Scaling Results}
\label{sec:result1}

Table~\ref{tab:pilot} compares FTTT with various baselines across four reasoning datasets. FTTT, both with and without self-reflected feedback, outperforms conventional test-time scaling methods on average. This success is partially explained by the findings of \citet{ye2024physicslanguagemodels22}, which show that training with error-correction data enhances reasoning capabilities and models barely retry during inference. FTTT is also efficient. For instance, the inference time of \texttt{Llama-3.1-8B-Instruct} on GSM8K with a budget of 32 is 3 GPU hours for the best parallel sampling method (Best-of-N) and 20 GPU hours for the best sequential revision method (Self-Refine). In contrast, FTTT achieves inference times of approximately 3 GPU hours without self-reflected feedback and 4 GPU hours with self-reflected feedback.


Notably, self-reflected feedback does not always improve results. Its effectiveness appears to depend on the LLM's self-reflection ability. To test this, we computed the Spearman rank correlation between FTTT and Self-Refine, a self-reflection-based algorithm. The Spearman coefficient ($r = 0.8656$, $p \leq 0.05$) indicates a strong positive correlation, supporting our hypothesis. We also observe that Self-Consistency performs poorly on code tasks because sampled code snippets rarely match exactly, making majority voting akin to random selection.

Figure~\ref{fig:scale} illustrates performance for FTTT and baselines under varying budgets. FTTT consistently outperforms baselines, with greater gains under constrained budgets. In contrast, Revision and Self-Consistency do not scale well. Revision struggles with long-context reasoning due to length generalization issues~\cite{li-etal-2024-making}, while Self-Consistency fails to leverage feedback, often discarding correct answers during majority voting due to long-tailed distributions of correct answers~\cite{DBLP:journals/corr/abs-2407-21787}.



\subsection{Fine-Tuning Results}

\begin{table*}
\centering
\small
\begin{tabular}[t]{l|r|ccc|c}
\toprule
 \makecell[c]{\textbf{Method}} & \makecell[c]{\textbf{\#Param.}} & \makecell[c]{\textbf{MATH}} & \makecell[c]{\textbf{GSM8K}} & \makecell[c]{\textbf{MBPP}} & \makecell[c]{\textbf{Avg.}} \\
\midrule
\multicolumn{6}{c}{\texttt{Llama-3.1-8B-Instruct}} \\
\midrule
Adapter~\cite{DBLP:conf/icml/HoulsbyGJMLGAG19} & 134M & 0.5933$_{0.0151}$ & 0.7979$_{0.0056}$ & 0.2632$_{0.0058}$ & 0.5515 \\
(IA)$^3$~\cite{DBLP:conf/nips/LiuTMMHBR22} & 524K & 0.6187$_{0.0105}$ & 0.8929$_{0.0107}$ & 0.5685$_{0.0022}$ & 0.6934 \\
LoRA~\cite{DBLP:conf/iclr/HuSWALWWC22} & 1.7M & 0.6387$_{0.0136}$ & 0.9186$_{0.0037}$ & 0.5639$_{0.0242}$ & 0.7071 \\
LN-Tuning~\cite{DBLP:conf/iclr/ZhaoT0MX24} & 266K & 0.6280$_{0.0113}$ & 0.8899$_{0.0056}$ & 0.5748$_{0.0175}$ & 0.6976 \\
Full Fine-Tuning & 8B & 0.6027$_{0.0136}$ & 0.7722$_{0.0056}$ & 0.4034$_{0.0096}$ & 0.5928 \\
\midrule
\method{} & 439K & \textbf{0.7013}$_{0.0050}$ & \textbf{0.9246}$_{0.0056}$ & \textbf{0.6184}$_{0.0159}$ & 0.7481 \\
\toprule
\multicolumn{6}{c}{\texttt{Mistral-7B-Instruct-v0.3}} \\
\midrule
Adapter~\cite{DBLP:conf/icml/HoulsbyGJMLGAG19} & 134M & \textbf{0.5418}$_{0.0111}$ & 0.8264$_{0.0021}$ & 0.2763$_{0.0076}$ & 0.5482 \\
(IA)$^3$~\cite{DBLP:conf/nips/LiuTMMHBR22} & 524K & 0.5041$_{0.0056}$ & 0.8686$_{0.0060}$ & 0.4914$_{0.0185}$ & 0.6214 \\
LoRA~\cite{DBLP:conf/iclr/HuSWALWWC22} & 1.7M & 0.5117$_{0.0091}$ & 0.8686$_{0.0016}$ & 0.4968$_{0.0046}$ & 0.6257 \\
LN-Tuning~\cite{DBLP:conf/iclr/ZhaoT0MX24} & 266K & 0.4357$_{0.0115}$ & 0.8259$_{0.0051}$ & 0.4065$_{0.0095}$ & 0.5560 \\
Full Fine-Tuning & 7B & 0.5388$_{0.0157}$ & 0.7355$_{0.0016}$ & 0.2548$_{0.0095}$ & 0.5097 \\
\midrule
\method{} & 439K & 0.4891$_{0.0111}$ & \textbf{0.9003}$_{0.0039}$ & \textbf{0.5194}$_{0.0070}$ & 0.6363 \\
\bottomrule
\end{tabular}
\caption{Fine-tuning results on four datasets with a budget of 32. \#Param. denotes the number of trainable parameters. We report the mean of three runs with different random seeds and standard deviation in the subscript. \textbf{Bold} entries are the best results.}
\label{tab:meta}
\end{table*}

\input{figures/finetune}

We present the results of PEFT baselines and \method{} with a budget of 32 in Table~\ref{tab:meta}. Best-of-N is applied to PEFT baselines to exploit test-time feedback. HumanEval is excluded as it lacks a training set. Table~\ref{tab:meta} highlights the effectiveness of \method{}, outperforming all PEFT baselines by at least 2.58\% on average. \method{} is also parameter-efficient, with 439K trainable parameters that are comparable to the most lightweight PEFT method (LN-Tuning, 266K parameters), while surpassing the best PEFT method (LoRA, 1.7M parameters) with an order of magnitude fewer parameters.
However, \method{} shows suboptimal performance on MATH for \texttt{Mistral-7B-Instruct-v0.3}, which is consistent with other PEFT methods with few trainable parameters (e.g., (IA)$^3$, LoRA, LN-Tuning). This is likely due to \texttt{Mistral-7B-Instruct-v0.3}'s limited mathematical reasoning capabilities, requiring significant parameter updates to improve performance in this domain.

\method{} incurs negligible inference overhead. For example, on GSM8K with \texttt{Llama-3.1-8B-Instruct} and a budget of 32, the best test-time scaling baseline (FTTT) requires 4 GPU hours, whereas \method{} uses only 1.5 GPU hours, benefiting from shorter yet accurate predictions.


\begin{table*}[t!]
    \centering
    \small
    \begin{tabular}{p{0.52\textwidth}|p{0.42\textwidth}}
    \toprule
    \makecell*[{{p{0.52\textwidth}}}]{
    \textbf{Question:} Blake and Kelly are having a contest to see who can run the most in 15 minutes. They decide to do it on a football field that is 100 yards long. Blake runs back and forth 15 times. Kelly runs back and forth once, and then decides that she doesn't want to run next to Blake, so she starts to run to the 40-yard line and back. She does this 34 times. How much farther does the winner run than the loser?
    \\
    \rule[0.4em]{0.52\textwidth}{0.4pt}
    \\
    \textbf{LoRA:} Blake runs 15 * 100 * 2 = 3000 yards.\\Kelly runs 100 * 2 = 200 yards for each run to the 40-yard line and back.\\\org{Kelly runs 200 * 34 = 6800 yards.}\\Kelly runs 6800 - 3000 = 3800 yards more than Blake.\\The final answer is \red{3800 \xmark}.
    \\
    \rule[0.4em]{0.52\textwidth}{0.4pt}
    \\
    \textbf{\method{}:} Blake runs back and forth 15 times, and each time, he runs 100 yards twice because he runs to one end of the field and back. So, Blake's total running distance is 15 x 200 = 3000 yards.\\Kelly runs back and forth once, which is 200 yards, and then runs to the 40-yard line and back, which is 40 x 2 = 80 yards, and does this 34 times. \blue{Kelly's total running distance is 200 + (34 x 80) = 200 + 2720 = 2920 yards.}\\Therefore, Blake runs 3000 - 2920 = 80 yards farther than Kelly.\\The final answer is \gr{80 \cmark}.
    }
    &
    \makecell*[{{p{0.42\textwidth}}}]{
    \textbf{Question:} Josh decides to take up juggling to perform at the school talent show a month in the future. He starts off practicing juggling 3 balls, and slowly gets better adding 1 ball to his juggling act each week. After the end of the fourth week the talent show begins, but when Josh walks on stage he slips and drops three of his balls. 2 of them are caught by people in the crowd as they roll off the stage, but one gets lost completely since the auditorium is dark. With a sigh, Josh starts to juggle on stage with how many balls?
    \\
    \rule[0.4em]{0.42\textwidth}{0.4pt}
    \\
    \textbf{LoRA:} Josh started with 3 balls and added 1 ball each week for 4 weeks, so he had 3 + 4 = 7 balls.\\\org{He dropped 3 balls, but 2 were caught, so he lost 3 - 2 = 1 ball.}\\He was left with 7 - 1 = 6 balls.\\The final answer is \red{6 \xmark}.
    \\
    \rule[0.4em]{0.42\textwidth}{0.4pt}
    \\
    \textbf{\method{}:} Josh started with juggling 3 balls. After the end of the fourth week, Josh would have added 4 balls, making it a total of 3+4=7 balls.\\\blue{Josh had 3 balls left after 2 of the balls were caught by people in the crowd, so he was left with 7-3=4 balls.}\\The final answer is \gr{4 \cmark}.
    }
    \\
    \bottomrule
    \end{tabular}
    \caption{Two GSM8K examples with model predictions. \gr{\cmark} means correct and \red{\xmark} means incorrect. Sentences marked in \org{orange} are incorrect reasoning steps and those marked in \blue{blue} are the correct counterparts.}
    \label{tab:case}
\end{table*}

Finally, Figure~\ref{fig:tune-scale} examines the scaling behavior of PEFT baselines and \method{}. Initially, \method{} underperforms compared to other PEFT methods and FTTT, but it mostly achieves superior results when the budget exceeds 2. \method{}'s weaker performance with smaller budgets arises from its reliance on an initial attempt sampled from the raw LLM to initiate the process. This initial attempt often fails but is still counted as a valid attempt, making \method{} less competitive in low-budget settings.

\subsection{Analysis}

\paragraph{Ablation Study.}

Table~\ref{tab:ablation} presents an ablation study on the architecture design of \method{}. The results demonstrate that all components are essential, as removing any of them significantly degrades performance. Notably, normalization is the most critical component, as it makes the input gradient scales more stable.


\paragraph{Case Study.}

Table~\ref{tab:case} provides two examples on GSM8K where the leading PEFT method, LoRA, fails, but \method{} succeeds using \texttt{Llama-3.1-8B-Instruct}. These examples highlight \method{}'s superior ability to correctly interpret and reason through questions, unlike LoRA.

\begin{table}
\centering
\small
\setlength{\tabcolsep}{5pt}
\begin{tabular}[t]{l|cc}
\toprule
\makecell[c]{\textbf{Method}} & \makecell[c]{\textbf{\#Params.}} & \makecell[c]{\textbf{Pass@1}} \\
\midrule
\textbf{\method{}} & 439K & 0.2600 \\
+ w/o Dropout & 439K & 0.2243 \\
+ w/o Norm & 439K & 0.2150 \\
+ Not Share $\theta_1$ \& $\theta_2$ & 878K & 0.2243 \\
\bottomrule
\end{tabular}
\caption{The ablation study of \method{}. We report results of \texttt{Llama-3.1-8B-Instruct} on MBPP with a budget of 2.}
\label{tab:ablation}
\end{table}


\section{Related Work}

\paragraph{Learning from Feedback.}

Other than the heuristic binary feedback studied in this work, prior research has explored feedback from various sources, such as humans~\cite{DBLP:conf/nips/Ouyang0JAWMZASR22}, other models~\cite{yang-etal-2022-re3}, tools~\cite{DBLP:conf/nips/SchickDDRLHZCS23}, and knowledge bases~\cite{gao-etal-2023-rarr}. This paper focuses on demonstrating the effectiveness of the proposed method and other feedback types are beyond the scope of this paper.

\paragraph{Test-Time Training.}

Test-Time Training (TTT) has shown success in the image modality by addressing distribution shifts and enhancing model capacity through self-supervised fine-tuning on each test case~\cite{DBLP:conf/icml/SunWLMEH20,DBLP:conf/nips/LiuKDBMA21,DBLP:journals/corr/abs-2310-13807}. Recent studies have extended TTT to the text modality~\cite{DBLP:conf/iclr/Hardt024,DBLP:journals/corr/abs-2401-11504}. The most relevant work, by~\citet{akyürek2024surprisingeffectivenesstesttimetraining}, uses TTT to enhance the reasoning ability of LLMs. However, their method relies heavily on human scaffolding for self-supervision and does not generalize beyond ARC-AGI~\cite{chollet2019measureintelligence}. In contrast, FTTT is generally applicable.

\paragraph{Learning to Optimize.}

Learning to Optimize (L2O) trains a network to act as an optimizer for another network~\cite{DBLP:journals/jmlr/ChenCCH0WY22}. Early approaches used reinforcement learning to train such optimizers~\cite{DBLP:conf/iclr/LiM17,DBLP:conf/icml/ChenHCDLBF17}, while recent work focuses on discovering analytical white-box optimizers~\cite{DBLP:conf/icml/BelloZVL17,DBLP:conf/nips/ChenLHRW0DLHLL23}. The most relevant work, MEND~\cite{DBLP:conf/iclr/MitchellLBFM22}, trains a network to predict weight updates from training gradients. \method{} builds on this idea, extending it to learn from test-time feedback with a distinct architecture.

\section{Conclusion}

In this paper, we propose a novel paradigm that leverages optimization to address the challenge of exploiting test-time feedback, resulting in improved scaling performance. We further present a learnable test-time optimizer, \method{}, which surpasses various PEFT baselines. Both FTTT and \method{} are efficient in terms of speed and trainable parameter count.

\section*{Limitations}

The current evaluation setting limits FTTT's potential by providing only binary feedback (i.e., correct or incorrect) for each attempt. However, developing complex reasoning environments with rich feedback is beyond the scope of this work. Additionally, while continuous feedback, such as that from reward models~\cite{DBLP:journals/corr/abs-2409-12122}, has been extensively studied, it is not examined here. Our method can be adapted to continuous feedback with minimal modifications, such as using REINFORCE~\cite{DBLP:journals/ml/Williams92}. For coherence, we leave this exploration to future work.

\section*{Acknowledgments}

This work is supported by National Key R\&D Program of China (Project No. 2022ZD0161200, 2022ZD0161201). This work is also supported by the Research Grants Council of the HKSAR - Early Career Scheme (Grant No. 24200223) and Hong Kong Innovation and Technology Commission Project No. ITS/228/22FP. It is also partially supported by RGC Senior Research Fellow Scheme Ref. No. SRFS2425-4S03.

\bibliography{custom}

\appendix

\section{Hyperparameter Settings}
\label{app:setting}

Below is the detailed configurations of different PEFT methods:
\begin{itemize}[noitemsep, nolistsep]
    \item \textbf{Adapter} uses a learning rate of 1e-4 and the reduction factor of the bottleneck is 16.
    \item \textbf{(IA)$^3$} uses a learning rate of 5e-5.
    \item \textbf{LoRA} uses a learning rate of 2e-5. We only apply LoRA to the query and value projections in the last 8 layers, with a rank of 16 and a dropout ratio of 0.05.
    \item \textbf{LN-Tuning} uses a learning rate of 4e-4.
    \item \textbf{Full Fine-Tuning} uses a learning rate of 1e-5.
\end{itemize}
The number of training epochs is 100, 10, and 3 for MBPP, GSM8K, and MATH, respectively. We use the Adam optimizer with a batch size of 20 for all methods in all datasets, including \method{}. We use the checkpoint of the last epoch for inference.

For \method{}, we sample 10 attempts for each training example together with the raw question to construct the model input. We employ nucleus sampling~\cite{DBLP:conf/iclr/HoltzmanBDFC20} with a temperature of 0.6 and $p=0.95$ to generate attempts. The number of training epochs for MBPP, GSM8K, and MATH is set to 10, 3, and 3, respectively. The learning rate is 1e-5. In inference, we sample an attempt using the same hyperparameters as in data generation before applying \method{} to mitigate the train-test discrepancy. We alternate between sampling attempts from the raw LLM and predicting refined attempts from sampled attempts when scaling \method{} with more budgets.

\begin{figure}[t!]
    \centering
    \makeatletter
    \let\ref\@refstar
    \ref{outsidelegend}
    \makeatother
    \\
    \begin{tikzpicture}
        \begin{axis}[
            width=0.49\textwidth,height=0.305\textwidth,
            legend cell align={left},
            legend pos=north west,
            enlargelimits=0.07,
            legend style={
                font=\small,
                draw=none,
                column sep=3pt,
                legend columns=3,
                cells={align=left},
            },
            legend to name=outsidelegend,
            yticklabel style={/pgf/number format/fixed,/pgf/number format/precision=1},
            ylabel={Loss},
            ylabel near ticks,
            xlabel={Training Step},
            xlabel near ticks,
            xmajorgrids=true,
            ymajorgrids=true,
            grid style=dashed,
            xtick={0,300,600,...,1800},
            xmin=0,xmax=1800,
            every tick label/.append style={font=\small},
            label style={font=\small},
            ylabel style={yshift=0pt},
        ]
            \addplot [lyygreen,thick] coordinates {
                (18,0.9542386468081551) (37,0.5520389169295203) (56,0.4238962999639664) (74,0.31498534666663186) (93,0.20807345793208973) (112,0.127919691769197) (130,0.08092199927345317) (149,0.06297820392139455) (168,0.07112695826566155) (186,0.03355343456574302) (205,0.016616783040092592) (224,0.008384858860689052) (243,0.005146607358187915) (261,0.003478074775022619) (280,0.00248216546793035) (299,0.002202527248923154) (317,0.002041674552754285) (336,0.0019245354886998467) (355,0.0018405662500922055) (373,0.001759648960541914) (392,0.0016819466562832102) (411,0.0016135283332457517) (430,0.0015653531181621042) (448,0.0015081924231932125) (467,0.001445999598120623) (486,0.0013911574282110693) (504,0.0013167127409083321) (523,0.0012651115337157632) (542,0.001218417908418625) (560,0.0011538924539790435) (579,0.001106366435474253) (598,0.0010715656898875926) (617,0.0010303758523043465) (635,0.0010039036605447372) (654,0.0009634042965537086) (673,0.0009328204998995531) (691,0.0009004280847661636) (710,0.0008720354759756894) (729,0.0008419131372064193) (747,0.0008228059439735617) (766,0.0007982842903086208) (785,0.0007831387341341232) (804,0.0007632690317490522) (822,0.0007532368050539557) (841,0.0007376152085747949) (860,0.0007285565775345991) (878,0.0007188228521755035) (897,0.0007141477124576263) (916,0.0007039775504147943) (934,0.0006932028952766867) (953,0.0006887623492409201) (972,0.0006867324445336898) (991,0.000679134446031907) (1009,0.0006726439145797077) (1028,0.0006716209936906947) (1047,0.0006676370447332209) (1065,0.0006629037745496169) (1084,0.0006605592720648821) (1103,0.0006571137570442363) (1121,0.0006561235868357082) (1140,0.000653361015141329) (1159,0.0006519822194614512) (1178,0.0006511643688309001) (1196,0.0006484827972988394) (1215,0.0006503154807549747) (1234,0.0006430800107711139) (1252,0.0006442652905688566) (1271,0.0006448444835642442) (1290,0.0006445447072625798) (1308,0.0006462373437090991) (1327,0.0006434422603902969) (1346,0.0006415263854246089) (1365,0.0006409153182876301) (1383,0.0006416683448827203) (1402,0.0006419450921170852) (1421,0.000641917162400516) (1439,0.0006420994426477402) (1458,0.0006397050571314153) (1477,0.0006416312513504436) (1495,0.000642530620098114) (1514,0.0006420875296873204) (1533,0.0006416542405750662) (1552,0.0006424973116201512) (1570,0.0006398589693926234) (1589,0.0006407788969616202) (1608,0.0006401856395012555) (1626,0.0006431720473549583) (1645,0.000642706405988989) (1664,0.0006396194750612432) (1682,0.0006407230375284817) (1701,0.0006428205951012392) (1720,0.0006404073639986987) (1739,0.0006408571081365494) (1757,0.0006415600126439875) (1776,0.0006425462782701706) (1795,0.0006406192871976027) (1813,0.0006418759252298325) (1832,0.0006422056233819156) (1851,0.000639415679768445)
            };\addlegendentry{Adapter}
            \addplot [lyyred,thick] coordinates {
                (18,1.7554693375041779) (37,1.7554690111129678) (56,1.6671127890520554) (74,1.3909660788143383) (93,1.2040278796843666) (112,1.1514428286628926) (130,1.1214643672188336) (149,1.0819559352283172) (168,1.0317342013598763) (186,0.9658487901330631) (205,0.9089434618618399) (224,0.8725025360597009) (243,0.8443474591097092) (261,0.8206350561131768) (280,0.8000459722019134) (299,0.7830206723136698) (317,0.7678534359855448) (336,0.7537633722478693) (355,0.7415858794023646) (373,0.7313682841744652) (392,0.7229792141021891) (411,0.7152821749926889) (430,0.7096122700900318) (448,0.7051578460530163) (467,0.7000796884138953) (486,0.6951528131005599) (504,0.6910648447944518) (523,0.6869377095431568) (542,0.6828643472436915) (560,0.6794290797595672) (579,0.6760967479032629) (598,0.6728261447845296) (617,0.6700128565497577) (635,0.6668298894708807) (654,0.6640657639120989) (673,0.660881776860691) (691,0.6580245890081885) (710,0.6557259789125167) (729,0.6528709534017797) (747,0.6509262737743358) (766,0.6489781670391879) (785,0.6466813112962693) (804,0.6445841661749039) (822,0.6421717271447819) (841,0.6403087677165149) (860,0.6384705324223973) (878,0.6372795716963987) (897,0.6354558608111214) (916,0.6336539365391043) (934,0.6318355295109876) (953,0.6307882625151445) (972,0.6289875214112634) (991,0.6277765692236589) (1009,0.6264372228938628) (1028,0.6255888505415483) (1047,0.6242508913743943) (1065,0.6225483532257896) (1084,0.6220851633000501) (1103,0.6209558497138202) (1121,0.6195210747540316) (1140,0.6190990917185412) (1159,0.6177656204305231) (1178,0.6176741900928517) (1196,0.616593019210081) (1215,0.6160685493346841) (1234,0.6145311651382854) (1252,0.6142568333263703) (1271,0.6139439873516879) (1290,0.6132310265525777) (1308,0.6130044294551095) (1327,0.6118666704963235) (1346,0.6116243658218792) (1365,0.611180718569832) (1383,0.6098323964817639) (1402,0.610548294801763) (1421,0.6096148975392713) (1439,0.6094170228682737) (1458,0.6090495471648354) (1477,0.6085322170971549) (1495,0.608281018262241) (1514,0.6080791860978234) (1533,0.608300683332637) (1552,0.6071604356408756) (1570,0.6071092738187249) (1589,0.6073427659305022) (1608,0.6067430628812249) (1626,0.6072237147366937) (1645,0.6066001035312918) (1664,0.6063325851358832) (1682,0.6067287424668909) (1701,0.6061174525296624) (1720,0.6058341857583766) (1739,0.6068142569638828) (1757,0.6068840230849974) (1776,0.6060360179227942) (1795,0.6062926022126713) (1813,0.6064312368790734) (1832,0.6060196575633983) (1851,0.6058715983508105)
            };\addlegendentry{(IA)$^3$}
            \addplot [purple,thick] coordinates {
                (18,1.7460645379867146) (37,1.6749987270742814) (56,1.5018238740808822) (74,1.287221082391586) (93,1.1221475040211397) (112,1.0266087169953209) (130,0.945939579112007) (149,0.8864494772518382) (168,0.8463257978306734) (186,0.8140307747743984) (205,0.7884622665650067) (224,0.7681764816855364) (243,0.7527012952508774) (261,0.7380089581331467) (280,0.7252134435317096) (299,0.7137992981283422) (317,0.7028738419639873) (336,0.6929551394865474) (355,0.6843065884024064) (373,0.6767618515912224) (392,0.6690995323466745) (411,0.661957480690696) (430,0.6556243896484375) (448,0.6491489920386656) (467,0.6434260893633021) (486,0.6374624731706425) (504,0.6320127191390583) (523,0.6268128583775484) (542,0.6222668000083557) (560,0.6169184312463444) (579,0.6122868175812584) (598,0.6076586718227774) (617,0.6036604203004888) (635,0.600121115618211) (654,0.5958626895027365) (673,0.5917645214713193) (691,0.5886610306520513) (710,0.5850229518298797) (729,0.5812995625052223) (747,0.5782756703422669) (766,0.5748209417822526) (785,0.5722173191009359) (804,0.5691095035981367) (822,0.5661965028487425) (841,0.5636940206435912) (860,0.561016184760925) (878,0.5578182444852942) (897,0.5554530913817054) (916,0.5529910816865808) (934,0.5511980107761322) (953,0.5486578303862383) (972,0.5464666845964238) (991,0.5442242137888537) (1009,0.5425344048974349) (1028,0.54019544468844) (1047,0.538796123973826) (1065,0.5367465503713026) (1084,0.5351155734954671) (1103,0.5332509005133481) (1121,0.5319006608769218) (1140,0.5303059542242856) (1159,0.5286179404845213) (1178,0.5275454597677139) (1196,0.5262143140170663) (1215,0.5246511857139873) (1234,0.5233735273228609) (1252,0.5224568576098764) (1271,0.5213587959819936) (1290,0.5200261406720004) (1308,0.5189899709772936) (1327,0.5183626062729779) (1346,0.5171647913315717) (1365,0.5164212313565341) (1383,0.5156473985967789) (1402,0.5152953040791067) (1421,0.5143313892384901) (1439,0.5137826032179562) (1458,0.513216355267693) (1477,0.5127909043255974) (1495,0.5123478282581676) (1514,0.5116932506867271) (1533,0.5114353200331091) (1552,0.5112253280884442) (1570,0.5106553266392672) (1589,0.5103230195886949) (1608,0.5098596664673505) (1626,0.5098760268267464) (1645,0.5095043896353818) (1664,0.5093571871997201) (1682,0.5092372792289857) (1701,0.5089196597828585) (1720,0.5090758379767922) (1739,0.5087888177065926) (1757,0.5087547914229612) (1776,0.5086430840313754) (1795,0.508950258958786) (1813,0.5085186065837024) (1832,0.5089081544927098) (1851,0.508757851340554)
            };\addlegendentry{LoRA}
            \addplot [brown,thick] coordinates {
                (18,2.3725053919827874) (37,2.3004018202184993) (56,2.0452322730406083) (74,1.795553605186748) (93,1.7523658466849097) (112,1.7370799671519885) (130,1.7210244571461397) (149,1.7080856568035594) (168,1.695120092381768) (186,1.683810310567764) (205,1.6726178663937166) (224,1.6612790357620322) (243,1.6503354648855282) (261,1.6383009313899566) (280,1.6252735158338905) (299,1.6175734576056986) (317,1.608700349369151) (336,1.6000119785574032) (355,1.5917284960415274) (373,1.5839610380284928) (392,1.5793820957449032) (411,1.5761300969251337) (430,1.5709703414835394) (448,1.5677116516439673) (467,1.563000357724766) (486,1.5590588574740976) (504,1.5559795196043615) (523,1.552694065685578) (542,1.549816274387951) (560,1.5476021996156417) (579,1.5457102729674967) (598,1.544226172136113) (617,1.542416496072861) (635,1.5417726894113468) (654,1.5407579391397894) (673,1.5407672412892714) (691,1.5389446727732286) (710,1.538028003060244) (729,1.5376559170809658) (747,1.536358838412851) (766,1.5359874052159928) (785,1.5349369151069518) (804,1.534805379449365) (822,1.5347010974578041) (841,1.5344897591493984) (860,1.5336284127464905) (878,1.5337803478546959) (897,1.5331533503404913) (916,1.5335380023813503) (934,1.5332475142045454) (953,1.5321281555502173) (972,1.5318616571273396) (991,1.5317852815842246) (1009,1.5308110038226939) (1028,1.53128068077373) (1047,1.531209690685578) (1065,1.5310257692388034) (1084,1.5308366255326704) (1103,1.5304410393862802) (1121,1.530734301888369) (1140,1.5304116641773897) (1159,1.5305684951537433) (1178,1.530960164605615) (1196,1.5311132420830549) (1215,1.5310673841180649) (1234,1.5306465026529077) (1252,1.5311165059951537) (1271,1.5307906043720756) (1290,1.5305247587316178) (1308,1.531179173107453) (1327,1.5314189074511197) (1346,1.5305250851228276) (1365,1.530712270481701) (1383,1.5308677958932153) (1402,1.5313423687123997) (1421,1.5311209122764873) (1439,1.5308790563899566) (1458,1.530742788059826) (1477,1.530255159592246) (1495,1.5307713472906919) (1514,1.530859472917363) (1533,1.5309211608560327) (1552,1.5306755514705883) (1570,1.5309064732515876) (1589,1.530676204253008) (1608,1.5310190782190007) (1626,1.5310983912830047) (1645,1.5311111205401906) (1664,1.5309555951286764) (1682,1.5305921585164606) (1701,1.5308467436601771) (1720,1.5307091697652073) (1739,1.5309978627903575) (1757,1.5306912182486632) (1776,1.5306607006705382) (1795,1.5308155732996322) (1813,1.530880525150401) (1832,1.5308860738009693) (1851,1.5310660785532253)
            };\addlegendentry{LN-Tuning}
            \addplot [lyyblue,thick] coordinates {
                (10,1.9096006155014038) (20,1.7125850915908813) (30,1.614748239517212) (40,1.5039660930633545) (50,1.5283615589141846) (60,1.466806173324585) (70,1.439060926437378) (80,1.3981423377990723) (90,1.4263986349105835) (100,1.4246621131896973) (110,1.3155015707015991) (120,1.38673996925354) (130,1.4046558141708374) (140,1.3570278882980347) (150,1.2971023321151733) (160,1.3223206996917725) (170,1.3192830085754395) (180,1.303603172302246) (191,1.2668116092681885) (201,1.174188494682312) (211,1.2061415910720825) (221,1.1700122356414795) (231,1.1616195440292358) (241,1.1596542596817017) (251,1.1091374158859253) (261,1.0903400182724) (271,1.0842281579971313) (281,1.052575707435608) (291,1.0578773021697998) (301,1.0312350988388062) (311,1.022706389427185) (321,0.9624451398849487) (331,0.9611892700195312) (341,0.9452405571937561) (351,0.9436296224594116) (361,0.914760947227478) (371,0.9535216689109802) (382,0.9140812158584595) (392,0.8824102878570557) (402,0.8976102471351624) (412,0.9014500379562378) (422,0.9064497351646423) (432,0.8495674729347229) (442,0.8406484723091125) (452,0.8989713788032532) (462,0.8461062908172607) (472,0.8367741107940674) (482,0.8454121947288513) (492,0.8448436260223389) (502,0.7907944321632385) (512,0.8239151239395142) (522,0.8640180230140686) (532,0.8193634748458862) (542,0.8770815134048462) (552,0.8740203976631165) (563,0.8575403094291687) (573,0.8226989507675171) (583,0.8681228160858154) (593,0.8066485524177551) (603,0.7744750380516052) (613,0.8078705072402954) (623,0.8055357933044434) (633,0.8131847977638245) (643,0.8110240697860718) (653,0.8219190239906311) (663,0.8297211527824402) (673,0.7941579222679138) (683,0.7894881367683411) (693,0.7672296762466431) (703,0.8334339261054993) (713,0.8101662993431091) (723,0.7439648509025574) (733,0.7701679468154907) (743,0.8044629096984863) (754,0.796661376953125) (764,0.7508822083473206) (774,0.7835989594459534) (784,0.8221855163574219) (794,0.8256995677947998) (804,0.7757627367973328) (814,0.7768563628196716) (824,0.7712129950523376) (834,0.786733865737915) (844,0.757821798324585) (854,0.7601490616798401) (864,0.7479444742202759) (874,0.7515830993652344) (884,0.7867664098739624) (894,0.7614544630050659) (904,0.752370297908783) (914,0.7334328293800354) (924,0.7403179407119751) (934,0.7982839941978455) (945,0.74163419008255) (955,0.726938784122467) (965,0.7719147205352783) (975,0.7532569766044617) (985,0.7286401987075806) (995,0.7569704651832581) (1005,0.7599449753761292) (1015,0.7469912767410278) (1025,0.752652108669281) (1035,0.7397856116294861) (1045,0.7200146317481995) (1055,0.707065224647522) (1065,0.7467519640922546) (1075,0.7649410963058472) (1085,0.7600623965263367) (1095,0.7756914496421814) (1105,0.7668757438659668) (1115,0.7188001275062561) (1126,0.7643749713897705) (1136,0.7547743320465088) (1146,0.7253299951553345) (1156,0.7218968272209167) (1166,0.735359251499176) (1176,0.7339749336242676) (1186,0.7134099006652832) (1196,0.7409310340881348) (1206,0.7373772263526917) (1216,0.7134911417961121) (1226,0.7231583595275879) (1236,0.7214823961257935) (1246,0.7427322864532471) (1256,0.7217097282409668) (1266,0.7022026777267456) (1276,0.7406946420669556) (1286,0.6812798380851746) (1296,0.7050934433937073) (1306,0.7238278985023499) (1317,0.7071363925933838) (1327,0.727793276309967) (1337,0.7169796228408813) (1347,0.7078614830970764) (1357,0.7413643598556519) (1367,0.7039703130722046) (1377,0.6938318610191345) (1387,0.6979725360870361) (1397,0.7060480713844299) (1407,0.6890267729759216) (1417,0.711628258228302) (1427,0.7204755544662476) (1437,0.7359045147895813) (1447,0.6808397173881531) (1457,0.6987863779067993) (1467,0.7334898114204407) (1477,0.7083642482757568) (1487,0.7080873847007751) (1498,0.693020761013031) (1508,0.6989137530326843) (1518,0.6913502216339111) (1528,0.6560222506523132) (1538,0.6944241523742676) (1548,0.7126632928848267) (1558,0.6960278153419495) (1568,0.6849088072776794) (1578,0.6883605122566223) (1588,0.6884991526603699) (1598,0.7009050250053406) (1608,0.7208908796310425) (1618,0.6864383816719055) (1628,0.7148692607879639) (1638,0.6848537921905518) (1648,0.6672382354736328) (1658,0.6999152302742004) (1668,0.6823928356170654) (1678,0.6755030155181885) (1689,0.6863232254981995) (1699,0.6999881863594055) (1709,0.7179044485092163) (1719,0.6620104908943176) (1729,0.6855274438858032) (1739,0.6795668005943298) (1749,0.6901704668998718) (1759,0.6914386749267578) (1769,0.6787542700767517) (1779,0.7018836736679077) (1789,0.6358839273452759) (1799,0.6689658164978027) (1809,0.6930421590805054) (1819,0.6745243072509766) (1829,0.6645663380622864) (1839,0.6246889233589172) (1849,0.678966224193573) (1859,0.6651533246040344) (1869,0.6703926920890808)
            };\addlegendentry{\method{}}
        \end{axis}
    \end{tikzpicture}
    \caption{The training curves of PEFT methods when fine-tuning \texttt{Llama-3.1-8B-Instruct} on MBPP.}
    \label{fig:learning}
\end{figure}

\section{Prompts}
\label{sec:prompt}

Below is the reflection generation prompts $P$ for \texttt{Llama-3.1-8B-Instruct} and \texttt{Mistral-7B-Instruct-v0.3}:

\patchcmd{\quote}{\rightmargin}{\leftmargin 15pt \rightmargin}{}{}
\begin{quote}
\small 
\begin{tcolorbox}[breakable, colback=white, colbacktitle=blue!5!white, colframe=black, boxrule=1pt, title={\textcolor{black}{\textbf{Llama-3.1-8B-Instruct}}}]
\textbf{User:} Solve the following math problem $\ldots$\\
\textbf{Assistant:} $\ldots$ the final answer is: $\ldots$\\
\textbf{User:} Your answer is incorrect. \textcolor{gray}{Please carefully check the solution and summarize all mistakes in short. Do NOT provide the corrected solution. Do NOT say ``my solution''.}\\
\textbf{Assistant:} \textcolor{royalblue}{Here is the summary of the mistakes in the previous solution $\ldots$}
\end{tcolorbox}
\end{quote}
\patchcmd{\quote}{\rightmargin}{\leftmargin 26pt \rightmargin}{}{}

\patchcmd{\quote}{\rightmargin}{\leftmargin 15pt \rightmargin}{}{}
\begin{quote}
\small 
\begin{tcolorbox}[breakable, colback=white, colbacktitle=blue!5!white, colframe=black, boxrule=1pt, title={\textcolor{black}{\textbf{Mistral-7B-Instruct-v0.3}}}]
\textbf{User:} Solve the following math problem $\ldots$\\
\textbf{Assistant:} $\ldots$ the final answer is: $\ldots$\\
\textbf{User:} Your answer is incorrect. \textcolor{gray}{Carefully check the solution step-by-step and list all mistakes in short. MUST NOT provide the correct answer. Your response MUST be in the third person tone.}\\
\textbf{Assistant:} \textcolor{royalblue}{Here is the summary of the mistakes in the previous solution $\ldots$}
\end{tcolorbox}
\end{quote}
\patchcmd{\quote}{\rightmargin}{\leftmargin 26pt \rightmargin}{}{}
Sentences in \textcolor{gray}{gray} are the prompt $P$ and the one in \textcolor{royalblue}{blue} is the generated reflection.

\section{Training Curves}

Figure~\ref{fig:learning} is the training curves of various PEFT methods, including \method{}. We observe that Adapter shows a clear signal of overfitting, where it has a training loss close to 0, while its performance on the test set is the worst. All PEFT methods seem to converge smoothly. For \method{}, its training is not as stable as baselines, suggesting the difficulty of learning to optimize problem.

\section{Transferability}

\begin{table}
\centering
\small
\setlength{\tabcolsep}{5pt}
\begin{tabular}[t]{lcc}
\toprule
\makecell[c]{\textbf{Model}} & \makecell[c]{\textbf{GSM8K}\\(trained on MATH)} & \makecell[c]{\textbf{HumanEval}\\(trained on MBPP)} \\
\midrule
\multicolumn{3}{c}{\cellcolor{lightgray}\texttt{Llama-3.1-8B-Instruct}}\\
Raw Model & 0.8069 & 0.6728 \\
+ \method{} & \textbf{0.8914} & \textbf{0.7037} \\
\midrule
\multicolumn{3}{c}{\cellcolor{lightgray}\texttt{Mistral-7B-Instruct-v0.3}}\\
Raw Model & 0.7807 & 0.6500 \\
+ \method{} & \textbf{0.7921} & \textbf{0.7100} \\
\bottomrule
\end{tabular}
\caption{The cross-task transferability of \method{}.}
\label{tab:transfer}
\end{table}

While generalization and transfer are not the primary focus of this work, we demonstrate \method{}’s transferability by training it on one dataset (MBPP/MATH) and testing it on another (HumanEval/GSM8K). Table~\ref{tab:transfer} indicates that \method{} consistently improves baseline performance, even when tested on unseen datasets. This highlights \method{}'s potential for generalization and transferability.

\end{document}